\documentclass{article}

\usepackage{microtype}
\usepackage{graphicx}
\usepackage{subfigure}
\usepackage{booktabs} %

\usepackage{hyperref}

\usepackage{amsmath}
\usepackage{amsfonts}
\usepackage[algo2e]{algorithm2e}
\usepackage{wrapfig}
\usepackage{caption}
\usepackage{multirow}

\newcommand{\E}[0]{\mathbb{E}}

\makeatletter
\newcommand{\removelatexerror}{\let\@latex@error\@gobble}
\makeatother
\usepackage[accepted]{icml2021}

\icmltitlerunning{Discriminator Augmented Model-Based Reinforcement Learning}

\begin{document}

\twocolumn[
\icmltitle{Discriminator Augmented Model-Based Reinforcement Learning}

\icmlsetsymbol{equal}{*}

\begin{icmlauthorlist}
\icmlauthor{Behzad Haghgoo}{equal,stanford}
\icmlauthor{Allan Zhou}{equal,stanford}
\icmlauthor{Archit Sharma}{stanford}
\icmlauthor{Chelsea Finn}{stanford}
\end{icmlauthorlist}

\icmlaffiliation{stanford}{Department of Computer Science, Stanford University, Stanford, USA}

\icmlcorrespondingauthor{Behzad Haghgoo}{bhaghgoo@stanford.edu}
\icmlcorrespondingauthor{Allan Zhou}{ayz@stanford.edu}

\icmlkeywords{Machine Learning, Reinforcement Learning}

\vskip 0.3in
]

\printAffiliationsAndNotice{\icmlEqualContribution} %

\begin{abstract}
By planning through a learned dynamics model, model-based reinforcement learning (MBRL) offers the prospect of good performance with little environment interaction. However, it is common in practice for the learned model to be inaccurate, impairing planning and leading to poor performance.
This paper aims to improve planning with an importance sampling framework that accounts and corrects for discrepancy between the true and learned dynamics. This framework also motivates an alternative objective for fitting the dynamics model: to minimize the variance of value estimation during planning. We derive and implement this objective, which encourages better prediction on trajectories with larger returns.
We observe empirically that our approach improves the performance of current MBRL algorithms on two stochastic control problems,
and provide a theoretical basis for our method. 
\end{abstract}

\section{Introduction}

Model free reinforcement learning methods have achieved good performance on a number of complex tasks, but usually require a large amount of data collected through environment interaction~\citep{mnih2015human,lillicrap2015continuous}. \textit{Model-based} reinforcement learning (MBRL) can potentially reduce these data requirements by fitting a model of the environment dynamics on a small dataset, then planning through the learned dynamics to produce a good policy. However, inaccurate models can severely impair planning performance. Recent techniques attempt to avoid \textit{compounding} model error by restricting the number of model unrolling steps \citep{janner2019trust,feinberg2018model}, but this creates a trade-off between planning performance and sample efficiency. More importantly, such approaches assume that the learned model mostly matches the true (one timestep) dynamics, and only address compounding error that arises over many timesteps. In this paper we show that there are a range of environments where the learned model can be inaccurate even on short time horizons (e.g., one timestep), so merely mitigating compounding error is insufficient. Given the evidence that even deep neural network models can struggle to learn complex, high dimensional distributions \citep{arora2018gans}, we expect inaccurate learned dynamics to be an increasingly important issue as we apply MBRL to harder and more realistic environments.

Consider a learned dynamics model as a generative model from which we sample transitions. Even if the model is inaccurate, some samples may be better than others. Our approach trains a discriminative model to assess the quality of sampled transitions during planning, and upweight or downweight value estimates computed from high and low quality samples, respectively. In section \ref{sec:bias_correction} we also show that this method is a form of likelihood-free importance sampling that, assuming the discriminator is an optimal classifier, produces unbiased value estimates for planning.
Since our discriminator can correct model error during planning, we are no longer restricted to fitting the dynamics using a maximum likelihood objective. Instead, we can learn \textit{biased} dynamics models with advantageous properties, such as reduced value estimation variance during planning. We derive and implement an objective function for learning these variance-minimizing models. Unlike maximum likelihood, our proposed objective incorporates trajectory return statistics into the model fitting process. We call our planning and model training framework \textbf{D}iscriminator \textbf{A}ugmented \textbf{M}BRL (or \textbf{DAM} for short).

To evaluate scenarios where the dynamics is difficult to learn, our experiments consider environments with stochastic and multi-modal dynamics.
We find that DAM significantly improves planning performance over existing MBRL algorithms in these environments.

\section{Preliminaries}
Consider a Markov Decision Process (MDP) 
$\mathcal{M} \equiv (\mathcal{S}, \mathcal{A}, p, r, \gamma, T)$ \citep{puterman1990markov}, where $\mathcal{S}$ denotes the state space, $\mathcal{A}$ denotes the action space, $p: \mathcal{S} \times \mathcal{A} \times \mathcal{S} \mapsto \mathbb{R}_{> 0}$ denotes the transition dynamics, $r: \mathcal{S} \times \mathcal{A} \mapsto \mathbb{R}$ denotes the reward function, $\gamma \in [0, 1]$ denotes the discount factor and $T$ denotes the maximum episode horizon. Our objective is to maximize $\E_{p(\tau)}[R(\tau)]$ where $R(\tau) = \sum_{t=1}^T\gamma^t r(s_t, a_t)$ denotes the discounted return for the trajectory $\tau =\{s_1, a_1, \ldots s_T\}$, which is sampled from
\begin{equation}
\label{eq:traj-dist}
p (\tau) = p(s_1)  \prod_{t = 1}^{T-1} p (s_{t + 1} |s_t, a_t) \pi (a_t |s_t).
\end{equation}
Here, $p : \mathcal{S} \mapsto \mathbb{R}_{> 0}$ denotes the initial state distribution, and $\pi(a_t \mid s_t)$ defines our controller which is optimized to maximized the expected discounted return. The overloaded notation $p$ is interpreted based on variable(s) for which the density is computed.

Since $p$ is can only be sampled sequentially in practice, MBRL \citep{langlois2019benchmarking, nagabandi2020deep}
learns an estimate of the transition dynamics $q: \mathcal{S} \times \mathcal{A} \times \mathcal{S} \mapsto \mathbb{R}_{> 0}$. We can also define ${q(\tau) = p(s_1)  \prod_{t = 1}^{T-1} q (s_{t + 1} |s_t, a_t) \pi (a_t |s_t)}$ analogous to $p(\tau)$. The model is generally learned by maximizing ${\E_{s', a, s \sim \mathcal{B}} \big[\log q(s' \mid s, a)\big]}$, where $\mathcal{B}$
denotes the dataset of transitions collected in $\mathcal{M}$.
The approximate transition dynamics can be used to learn a parametric controller $\pi$ or perform online planning using sampling based methods.

\section{Discriminator Augmented MBRL}
\label{sec:approach}
\begin{algorithm2e}[t]
\SetAlgoLined
\SetKwInOut{Input}{input}
\SetKwProg{Fn}{Function}{}{}
\Fn{\textsc{EstimateValue}}{
 \Input{$s_1$: Initial state}
 \Input{$\mathbf{a}=\{a_t\}_{t=1}^H$: Action sequence}
 $w \gets 1$ \tcp*{stores weights (Eq.~\ref{eq:improved-estimator})}
 $R \gets 0$\;
 \For{$t \gets 1$ \KwTo $H$}{
   $r_t \gets r(s_t,a_t)$\;
   $R \gets R + w\cdot r_t$\;
   $s_{t+1}\sim q(\cdot | s_t,a_t)$\;
   $w \gets w \cdot \sigma(\Tilde{\mathcal{D}}(s_t,a_t,s_{t+1}))$ \tcp*{Eq.~\ref{eq:timestep-weight}}
 }
 \KwResult{$R$}
 \vspace{1em}
}
\caption{\textsc{EstimateValue} produces an estimated return $R$ of some action sequence using a learned model $q$. Since $q$ may be biased with respect to the true dynamics, it uses a discriminator $\mathcal{D}$ to correct the estimate as described in Sec.~\ref{sec:bias_correction}.}
 \label{alg:est-value}
\end{algorithm2e}

In this work, we look at some of the fundamental components of model-based reinforcement learning. In Section~\ref{sec:bias_correction}, we discuss an often ignored discrepancy caused by sampling trajectories from learned dynamics models and discuss how this bias can be corrected using importance sampling. Then, in Section~\ref{sec:min_variance_model_learning}, we discuss how interpreting the model-based RL in an importance sampling framework suggests a novel objective for learning models for environment dynamics.

\subsection{Correcting the Sampling Bias in Model-Based RL}
\label{sec:bias_correction}
Consider our optimization objective $J(\pi) = \E_{p(\tau)}[R(\tau)]$. Using the approximate dynamics $q$ introduces a bias in our estimation of $J(\pi)$. Model-based RL methods generally use ${\Tilde{J}_q(\pi) = \E_{q(\tau)}[R(\tau)]}$ as a proxy for $J(\pi)$, where the trajectories are sampled from $q(\tau)$ instead of $p(\tau)$ without accounting for the resulting bias. In this work, we mitigate this bias by using the importance sampling framework:
\begin{align}
    \label{eq:original-estimator}
    \E_{p(\tau)}\left[R(\tau)\right] &= \E_{q(\tau)}\Big[\frac{p(\tau)}{q(\tau)}R(\tau)\Big] =\E_{q(\tau)}\Big[w(\tau)R(\tau)\Big]
\end{align}
where we have introduced the importance sampling correction $w(\tau)$ to correct for the bias introduced by sampling trajectories from $q(\tau)$. Simplifying $w(\tau)$, we get
\begin{equation}
    w(\tau) = \prod_{t=1}^{T-1}\frac{p(s_{t+1} \mid s_t, a_t)}{q(s_{t+1} \mid s_t, a_t)} = \prod_{t=1}^{T-1} w(s_{t+1}, a_t, s_t)
\end{equation}
By exploiting the MDP's temporal structure, we can obtain an improved importance sampling estimator that weights the per-timestep rewards instead of the trajectory return. Comparing to Eq.~\ref{eq:original-estimator}, we see that this estimator multiplies the reward at time $t$ by a weight that only depends on transitions leading up to it, and not transitions that occur after it:
\begin{equation}
    \label{eq:improved-estimator}
    \mathbb{E}_{p(\tau)}[R(\tau)] = \sum_{t=1}^T\mathbb{E}_q\left[\left(\prod_{m=1}^t w(s_m,a_m,s_{m+1})\right)r(s_t,a_t)\right]
\end{equation}
As stated earlier, we only have access to the samples from the transition dynamics and not the probabilities. Thus to compute $w(\tau)$ for $\tau \sim q$, we use techniques from density ratio estimation~\citep{tran2017hierarchical,grover2019bias, sugiyama2012density}. We setup a binary classification problem over $\mathcal{D}: \mathcal{S} \times \mathcal{A} \times \mathcal{S} \times \mathcal{Y} \mapsto [0, 1] $, where $\mathcal{Y} = \{0, 1\}$. Using the transition samples from the environment and the learned dynamics, we create a dataset of transitions $\{s', a, s, y\}$,
where $y=1$ for $(s', a, s) \sim p(s' \mid s, a)p(s, a)$ and $y=0$ for $(s', a, s) \sim q(s' \mid s, a)p(s, a)$. Effectively, we are training a discriminator to distinguish between real transitions in the environment and the transitions sampled from the learned model. Alg.~\ref{alg:train} provides the pseudocode for training the discriminator.

To connect the discriminator to the importance weights, consider the Bayes-optimal classifier for this problem:
\begin{align}
    \mathcal{D}(s', a, s) &= \frac{p(s' \mid s, a)p(s,a)}{p(s' \mid s, a)p(s,a) + q(s' \mid s, a)p(s, a)}\\
        &= \frac{p(s' \mid s, a) / q(s' \mid s, a)}{p(s' \mid s, a) / q(s' \mid s, a) + 1}\\
        &= \sigma\Big(\log \frac{p(s' \mid s, a)}{q(s' \mid s, a)}\Big) = \sigma\Big(\Tilde{\mathcal{D}}(s', a, s)\Big)
\end{align}
$\Tilde{\mathcal{D}}$ is the classifier logits, and we can recover the importance weight by exponentiating them:
\begin{equation}
\label{eq:timestep-weight}
w(s', a, s) = \frac{p(s', a, s)}{q(s', a, s)} = \exp\big(\Tilde{\mathcal{D}}(s', a, s)\big)
\end{equation}

For our transitions $\{s', a, s\}$, it is important to sample $s, a \sim p(s, a)$ (that is, real state-action pairs) so that the logits can be connected to the importance weights. Since, we only need samples to train the classifier, this allows us to estimate the importance weights without requiring probabilities from underlying transition dynamics. Alg.~\ref{alg:est-value} provides the pseudocode for estimating the discriminator-corrected return of an action sequence, which forms the subroutine for choosing actions in sampling based planning algorithms.

\begin{wrapfigure}{r}{0.4\linewidth}
    \captionsetup{format=plain}
    \centering
    \includegraphics[width=\linewidth]{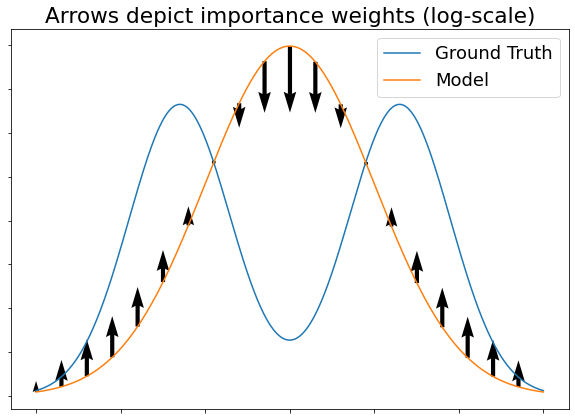}
    \caption{\small In this illustration, the blue curve depicts a ground truth density and orange a learned model. Importance sampling weights (represented by each arrow's magnitude in log-scale) can help us correct estimates under the incorrect model distribution.}
    \label{fig:importance}
\end{wrapfigure}
The benefit of this approach arises from the potential simplicity of the discriminative problem compared to the modelling the dynamics. The dynamics can be hard to model in the real world, as they often involve higher dimensional spaces (like RGB images) and can be multi-modal due to partial observability. The modelling and optimization challenges can often result in an incorrectly learned model, an example of which is shown in Fig~\ref{fig:importance}. On the other hand, discrimination can be an easier problem as fake and real samples can be well separated, especially in higher-dimensional spaces. This can improve the value estimation by down-weighting samples which do not fit with the real world samples. If the learned dynamics are indeed correct, the true and fake samples will be indistinguishable and the importance sampling weights will be close to unity, reducing the problem to conventional model-based RL.

\subsection{Minimum Variance Dynamics Model Fitting}
\label{sec:min_variance_model_learning}
\noindent The above discussion applies to any learned model, including the conventional approach of learning by maximizing the log-likelihood of transitions collected in the environment. The policy objective under $q$, that is $\hat{J}_q(\pi) = \frac{1}{N}\sum_{i=1}^N w(\tau_i) R(\tau_i)$, is an unbiased estimator of $J(\pi)$ $\forall q$ as long as the support of $p$ is a subset of support of $q$. However, from an optimization perspective, it is desirable to have an estimator with the minimum variance. Sampling based planning algorithms generate candidate action sequences and rank them according to their value function. Minimizing the variance of the value estimate of an action sequence increases the likelihood of the action sequences getting ranked correctly, and thus, the best action sequence getting chosen. Thus, it is favorable to reduce the variance of our estimate $\hat{J}_q$. Assuming that the $R(\tau) \geq 0$, it can be shown that
\begin{equation}
    q^*(\tau) = \frac{R(\tau)p(\tau)}{\mathbb{E}_{p(\tau)}\left[R(\tau)\right]}
\end{equation}
Here, $q^*$ denotes the sampling distribution which minimizes the variance of $\hat{J}_q$. While the denominator for the optimal sampling distribution is intractable, the important distinction here is that trajectories should be sampled in accordance to their density under the true trajectory distribution and the return accumulated by the trajectory. Using the $q^*$ as the target distribution, we can setup our minimization objective for model-based learning to be KL$\left(q^*(\tau) \mid \mid  q(\tau)\right)$. As we show in Appendix~\ref{appendix:model_gradient}, the gradient with respect to $q$ is:
\begin{align}
\label{eq:minvar-grad}
        \nabla_q \textrm{KL}&(q^*(\tau) \mid \mid q(\tau)) \nonumber \\&\propto - \sum_{t=1}^{T-1}\E_{p(\tau)}\left[R(\tau)\nabla_q \log q(s_{t+1} \mid s_t, a_t) \right]
\end{align}
Intuitively, this corresponds to upweighting the gradient for trajectories with higher return, forcing the dynamics model to fit better to transitions with higher return. In particular, this highlights that the conventional approach of learning dynamics does not represent the optimal approach in the importance sampling framework. The modified model training algorithm is described in Alg.~\ref{alg:train} in the appendix.

\subsection{Planning and Training Loop}
\noindent The previous two sections describe how to train the dynamics model, and how to train a discriminator that corrects the sampling bias from that learned model. Further, \textsc{EstimateValue} (Alg.~\ref{alg:est-value}) shows how to implement discriminator-corrected value estimation. We can plug \textsc{EstimateValue} as a subroutine into most sampling based planners. Once a planner is in place, we can use it execute actions in the real environment, collect more data, and re-train our model and discriminator. In Appendix~\ref{appendix:mbrl-loop} we discuss in detail the implementation of this model-based RL loop within our model and discriminator training framework.

\section{Experiments}
\begin{figure*}
    \centering
    \includegraphics[width=0.55\textwidth]{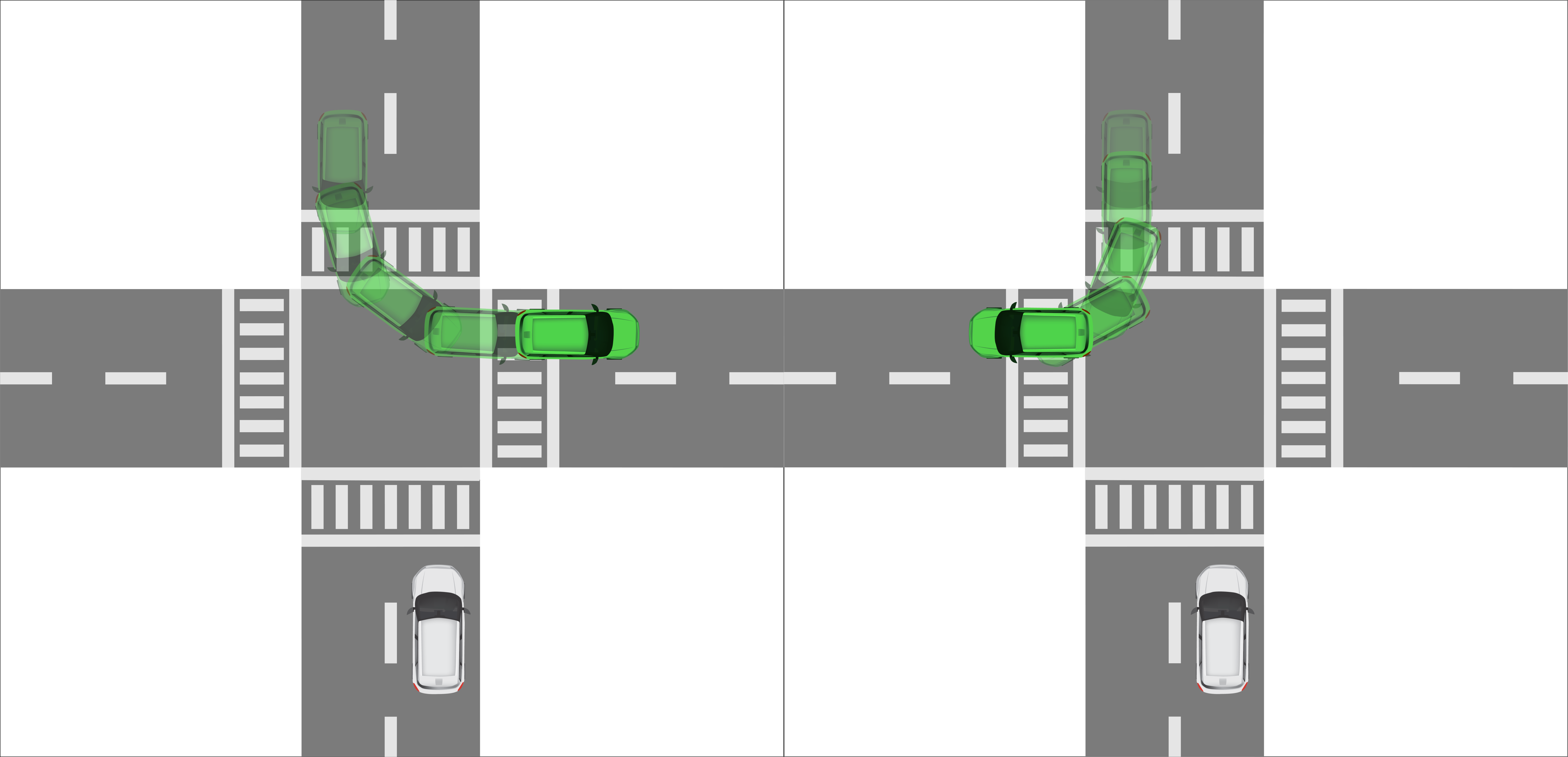}
    \includegraphics[width=0.35\textwidth]{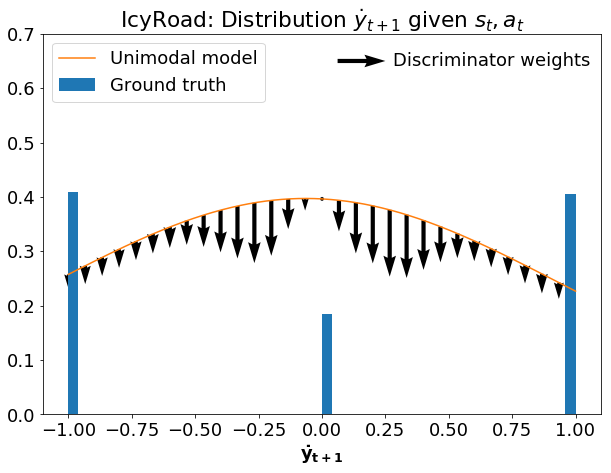}
    \caption{\textbf{Left}: Depiction of different scenarios in the Intersection environment, where the agent controls the silver car. Both cars are entering the intersection at the same time, but the agent does not know whether the oncoming green car will turn left (left panel) or turn right (right panel). If the agent drives into the intersection too fast, it may collide. \textbf{Right}: The learned (unimodal) model and ground truth distributions $p(\dot{y}_{s+1}|s_t,a_t)$, where $s_t=(2,0,2,0)$ and the action is to accelerate. After training with maximum likelihood, the unimodal model erroneously places large probability mass on unrealistic regions of the state space. The discriminator produces importance weights (black arrows, log scale) that downweight the model's samples in the unrealistic regions.}
    \label{fig:intersection-icyroad}
\end{figure*}
In this section we evaluate our proposed approach and compare with model-based baselines. We hypothesize that environments with multi-modal dynamics make model fitting difficult, which would pose problems for planning. Indeed, we observe that our proposed approach improves performance over standard model-based RL in such settings. 

We devise and implement two environments with stochastic, multi-modal dynamics: \textbf{IcyRoad} and \textbf{Intersection}. In \textbf{IcyRoad} the agent drives a car along an icy road, and above a threshold speed the car has a probability of swerving off the road in either direction (hence the multi-modality) and incurring a large negative reward. In \textbf{Intersection} the agent drives a car into an intersection. An oncoming car is also entering the intersection and will turn either left or right with equal probability--the outcome is unknown in advance to the agent. Fig.~\ref{fig:intersection-icyroad} illustrates the Intersection environment, and Appendix~\ref{appendix:envs} both environments in more detail.

For our baseline, we fit a dynamics model using the usual maximum likelihood loss and use a standard random shooting planner with uncorrected value estimates. In \textbf{DAM} we use both aspects of our method: we train the model using the minimum variance objective (Eq.~\ref{eq:minvar-grad}) and plan using the discriminator-corrected estimator (Eq.~\ref{eq:improved-estimator}). To disentangle the effects of each component of our method, we also evaluate variants that use only one aspect of our method, but not both. \textbf{DAM (disc only)} only does discriminator-corrected planning, while \textbf{DAM (minvar only)} only uses the minimum variance object.
Practitioners typically train models that output the parameters for a Gaussian distribution, which is unimodal. Since our environment dynamics are multimodal, it also makes sense to try fitting multimodal models such as mixture density networks \citep{bishop1994mixture}. We test each of our planning and training methods with both a mixture density network model and a standard Gaussian model, and display both results.

\begin{table*}[]
\centering
\begin{tabular}{|l|l|l|l|l|}
\hline
\multirow{2}{*}{Average return} & \multicolumn{2}{c|}{IcyRoad} & \multicolumn{2}{c|}{Intersection} \\
 & Unimodal & Multimodal & Unimodal & Multimodal \\
 \hline
Baseline & $0.71 \pm 1.72$ & $5.34 \pm 0.84$ & $62.67 \pm 22.10$ & $29.58 \pm 0.105$ \\
DAM (minvar only) & $0.86 \pm 0.79$ & $4.55 \pm 3.50$ & $31.40 \pm 1.111$ & $31.30 \pm 1.634$ \\
DAM (disc only) & $4.88 \pm 0.50$ & $3.30 \pm 2.75$ & $87.79\pm 15.34$ & $60.18 \pm 19.58$ \\
DAM & $2.32 \pm 1.76$ & $\mathbf{7.94 \pm 0.92}$ & $46.76\pm 9.543$ & $\mathbf{97.72\pm 5.127}$\\
\hline
\end{tabular}
\caption{\label{table:toy-results} Average return for each method in the IcyRoad and Intersection environments. Columns are split by whether they use a unimodal (Gaussian) or multimodal (mixture density network) learned dynamics model. Combining our method \textbf{DAM} with multimodal learned dynamics models usually performs best among different methods we evaluated. Results show mean and standard error across 3 runs (random seeds) for each hyperparameter setting.}
\end{table*}
Table~\ref{table:toy-results} depicts the results after 10 rounds of interleaving model (+ discriminator) training with data collection in the environment. We see that in both environments, \textbf{DAM} using multimodal dynamics models obtains the best results. \textbf{DAM (minvar only)}, which only uses the minimum variance objective without discriminator-corrected planning, is often \textit{worse} than the baseline. We hypothesize this occurs because the minimum variance objective increases model bias versus standard maximum likelihood training, which can be harmful without discriminator-corrected planning. When using unimodal models in either environment, we see that simply having a discriminator for planning already increases performance over the baseline. Fig.~\ref{fig:intersection-icyroad} depicts our analysis of what the trained model and discriminator are doing in IcyRoad: we see that the unimodal model is struggling to fit the true dynamics, while the discriminator is helping to correct the model error. Interestingly, adding the minimum variance objective to discriminator-corrected planning does \textit{not} help with unimodal models, but it \textit{does} help with multimodal models.

\section{Related Work}
Model-based RL has a rich history in control and robotics. Classically, the methods for model-based control have assumed access to underlying dynamics \citep{tassa2012synthesis, li2004iterative, mordatch2012discovery, lowrey2018plan} or make simplifying assumptions about the environment dynamics \citep{deisenroth2011pilco, deisenroth2013gaussian, levine2016end, kumar2016optimal}. While such simplifying assumptions can yield exact or efficient controllers, the models cannot faithfully represent the underlying dynamics, especially when considering high dimensional state spaces like RGB images. Recent work has focused on combining high-capacity function approximators such as neural networks with model-based reinforcement learning \citep{langlois2019benchmarking, nagabandi2018neural, nagabandi2020deep, chua2018deep, gal2016improving, sharma2019dynamics, hafner2019learning, nair2020goal}, which employ sampling-based planning methods \citep{williams2015model, nagabandi2018neural, rubinstein2013cross}. Concurrently, some recent work has integrated policy networks with deep models \citep{janner2019trust, wang2019exploring} in spirit of Dyna \citep{sutton1990integrated}. Neural networks allow more flexible representation of non-linear dynamics, allowing model-based RL to scale to complex high dimensional tasks \citep{kaiser2019model, nagabandi2020deep}. However, despite the incredible progress, none of these prior works account for the bias introduced by using learned models for planning.

Classically, the field of robust control has looked at designing controllers for with uncertain transition models \citep{bagnell2001solving, el2005robust}. More relevantly, \citep{abbeel2006using} incorporates bias correction for approximate transition models to learn better policies. However, they only focus on approximation error which arises due to the use of simple linear models. More recently, \citep{lambert2020objective} discusses the mismatch which arises between the training objectives and the control objectives, and how that discrepancy can lead to biases in model learning. In contrast to these works, our work focuses on the sampling bias introduced despite using flexible function approximators like neural networks. To some extent, our work also addresses the bias discussed in \citep{lambert2020objective}, as our objective encourages the transition model to fit better to trajectories with higher return.

Prior work has considered estimating value functions using transition models \citep{heess2015learning, feinberg2018model, buckman2018sample}. However, our work addresses the biased values generated by the use of learned transition models. To that extent, we rely on classifier-based density ratio estimation techniques \citep{sugiyama2012density}, which have been employed in generative modelling \citep{tran2017hierarchical, grover2019bias}. \citet{eysenbach2020off} use similar principles to correct for differences in dynamics when transferring experience between environments, where the classifier's corrections are realized by adjusting the reward appropriately before applying model-free RL. To the best of our knowledge, ours is first work which raises and addresses the issue of sampling bias in context of model-based RL.

\section{Conclusion}
We introduce a framework for mitigating the impact of model error on planning in model-based RL. Our framework trains discriminators to correct value estimation bias, while also introducing a novel model training objective for minimizing value estimation variance during planning. Empirically, we find that these modifications improve agent performance in environments with stochastic and multi-modal dynamics that pose a challenge for standard model fitting techniques. By improving planning in environments that are difficult to model properly, this framework is a step towards more efficient and performant reinforcement learning in complex, real world problems.

\bibliography{example_paper}
\bibliographystyle{icml2021}

\clearpage
\appendix
\section{Expanded Calculations}
\subsection{Minimum Variance Model Gradient}
\label{appendix:model_gradient}
Recall that under our assumptions the minimum variance trajectory distribution is:
\begin{equation}
    q^*(\tau) = \frac{R(\tau)p(\tau)}{\mathbb{E}_{p(\tau)}\left[R(\tau)\right]}
\end{equation}
The trajectory distribution under our learned model is $q(\tau)$, and the objective is to minimize $\text{KL}(q^\star(\tau) || q(\tau))$ by gradient based optimization. Then the gradient of the objective with respect to $q$ is:
\begin{align}
    \nabla_q \textrm{KL}(q^*(\tau) &\mid \mid q(\tau)) \\
     &= \nabla_q \int q^*(\tau) \log \frac{q^*(\tau)}{q(\tau)} d\tau\\
     &= - \int q^*(\tau) \nabla_q \log q(\tau) d\tau\\
     &= - \int p(\tau) \frac{R(\tau)}{\mathbb{E}_{p(\tau)}\left[R(\tau)\right]} \nabla_q \log q(\tau)\\
     &= - \E_{p(\tau)}\left[\frac{R(\tau)}{\mathbb{E}_{p(\tau)}\left[R(\tau)\right]} \nabla_q \log q(\tau)\right]\\
     &\propto - \E_p\left[R(\tau) \nabla_q \log q(\tau)\right]
\end{align}
The proportionality is because $\E_{p(\tau)}\left[R(\tau)\right]$ is a constant which can be be absorbed into the learning rate . Now using the definition of a trajectory distribution (Eq. \ref{eq:traj-dist}) we see:
\begin{equation}
    \nabla_q \log q(\tau) = \sum_{t=1}^{T-1} \nabla_q \log q(s_{t+1}|s_t,a_t)
\end{equation}
Giving us the result:
\begin{equation}
    \nabla_q \textrm{KL}(q^*(\tau) \mid \mid q(\tau)) \propto -\sum_{t=1}^{T-1}\mathbb{E}_p\left[R(\tau)\nabla_q \log q(s_{t+1}|s_t,a_t)\right]
\end{equation}

\section{Planning and training loop}
\label{appendix:mbrl-loop}
Here we expand on how model learning, value estimation, and planning fit together in a model-based RL loop. We can plug in \textsc{EstimateValue} (Alg.~\ref{alg:est-value}) as a subroutine that produces unbiased return estimates into many planning algorithms. For example, Alg.~\ref{alg:mbrl} defines \textsc{RSPlanner}, a simple random shooting planner. At each timestep, \textsc{RSPlanner} samples $K$ distinct sequences of actions, with each sequence containing $H$ (the planning horizon) sampled actions. Each individual action is sampled independently from a simple distribution $\pi$ over the action space, such as a uniform or Gaussian distribution. \textsc{RSPlanner} then uses \textsc{EstimateValue} to estimate a return for each action sequence, and selects the sequence with the highest estimated return. The planner then returns the first action in that sequence as the action for the current timestep. The planning process is repeated at the next timestep, and so on until the end of the episode.

Current approaches in model-based reinforcement learning tend to integrate planning and model training into an iterative loop: after training the model, they execute the planner in the real environment to collect some more data, and use this data to retrain the model. The pseudocode for \textsc{MBRL} in Alg.~\ref{alg:mbrl} shows how to do this with our framework. As \textsc{MBRL} uses \textsc{RSPlanner} to rollout more trajectories, it stores the observed transitions and retrains the model (\textsc{TrainModel}) and discriminator (\textsc{TrainDiscriminator}) on the new data. We define those model and discriminator training subroutines in Alg.~\ref{alg:train}. Crucially, the objective in \textsc{TrainModel} is return weighted as per Eq.~\ref{eq:minvar-grad}, which trains the model to minimize value estimate variance in planning.

\begin{figure*}
\begin{minipage}{0.49\textwidth}
\begingroup
\removelatexerror
\begin{algorithm2e}[H]
\SetAlgoLined
\SetKwInOut{Input}{input}
\SetKwProg{Fn}{Function}{}{}
\Fn{\textsc{TrainModel}}{
 \Input{Trajectory dataset $\mathcal{T}=\{\tau\}$}
 Initialize $q(\cdot|s,a)$ with random parameters\;
 \While{not converged}{
   $\tau=\{s_t,a_t\}_{t=1}^T\sim \mathcal{T}$\;
   \tcc{See Eq.~\ref{eq:minvar-grad}}
   $L=-\sum_t R(\tau)\log q(s_{t+1}|s_t,a_t)$\;
   Update $q$ by $\nabla L$
 }
 \KwResult{Trained model $q$}
 }
 \Fn{\textsc{TrainDiscriminator}}{
 \Input{Learned model $q(\cdot|s,a)$}
 \Input{Transition dataset $\mathcal{T}=\{(s,a,s')\}$}
 Initialize $\mathcal{D}$ with random parameters\;
 \While{not converged}{
   $(s,a,s')\sim \mathcal{T}$ \tcp*{Sample transition}
   $\Tilde{s}'\sim q(\cdot|s,a)$ \tcp*{Fake next state}
   \tcc{Classification loss}
   $L=-\log(\mathcal{D}(s,a,s')) - \log(1-\mathcal{D}(s,a,\Tilde{s}'))$
   Update $\mathcal{D}$ by $\nabla L$
 }
 \KwResult{Trained discriminator $\mathcal{D}$ (logits $\Tilde{\mathcal{D}}$)}
 }
 \caption{Training dynamics model and discriminator}
\label{alg:train}
\end{algorithm2e}
\endgroup
\end{minipage}
\begin{minipage}{0.49\textwidth}
\begingroup
\removelatexerror
\begin{algorithm2e}[H]
\SetAlgoLined
\SetKwProg{Fn}{Function}{}{}
\SetKwInOut{Input}{input}
\Fn{\textsc{RSPlanner}}{
 \Input{Current state $s_t$}
 \Input{Plan horizon $H$}
 \Input{Action sampling distribution $\pi$}
 \For{$i=1$ \KwTo $K$}{
 \tcc{Sample action sequences}
 $\mathbf{a}^{(i)} \gets \{a_1^{(i)},\cdots,a_H^{(i)}\}\stackrel{i.i.d}{\sim} \pi$\;
 }
 \tcc{Select best action sequence}
 $\mathbf{a} \gets \arg\max\limits_{\mathbf{a}^{(i)}} \mbox{\textsc{EstimateValue}}(s_t,\mathbf{a}^{(i)})$\;
 $a_t\gets a_1$\;
 \KwResult{$a_t$}
}
\Fn{\textsc{MBRL}}{
\Input{\textsc{Rollout}$(\cdot)$: Executes given planner in real environment, produces trajectory}
$\mathcal{T} \gets \{\}$\tcp*{Trajectory dataset}
\For{$j=1$ \KwTo $J$}{
$\tau \gets$ \textsc{Rollout}(\textsc{RSPlanner})\;
$\mathcal{T} \gets \mathcal{T} \cup \{\tau\}$\;
$q(\cdot|s,a) \gets $\textsc{TrainModel}$(\mathcal{T})$\;
$\mathcal{D} \gets $\textsc{TrainDiscriminator}$(q(\cdot|s,a), \mathcal{T})$\;
}
}
 \caption{Planning and MBRL Loop}
\label{alg:mbrl}
\end{algorithm2e}
\endgroup
\end{minipage}
\end{figure*}

\section{Environments}
\label{appendix:envs}
\noindent\textbf{IcyRoad:} The agent drives a car down a straight but icy road with extends in the $+x$ direction. If the car exceeds a threshold $x$-velocity ($\dot{x} > 2$), there is a high chance of swerving off the road in either the $+y$ or $-y$ directions, incurring a negative reward. The four dimensional observations give the car's position and velocity $s=(x, y, \dot{x}, \dot{y})$, and there are $3$ discrete actions for accelerating (increase $\dot{x}$ by $1$), decelerating (decrease $\dot{x}$ by $1$) and cruising ($\dot{x}$ stays the same). The velocities and states are updated using simple Euler integration, and the reward is $\dot{x}$ with bonuses for higher speeds and penalties for swerving:
\begin{equation}
    r(s,a)=\begin{cases}\dot{x}-6,\quad |y|>1\text{ (off road)}\\
    \dot{x} + 8,\quad |y|\leq 1,\dot{x}>2 \text{ (high speed bonus)}\\
    \dot{x} \quad \text{otherwise}\end{cases}
\end{equation}
The agent's car is always initialized with $(x,y,\dot{x},\dot{y})=(0,0,1,0)$, and each episode proceeds for 5 timesteps before terminating.

\noindent\textbf{Intersection:} Here the agent again controls a car, this time approaching an intersection with an oncoming car entering the intersection at the same time (Fig.~\ref{fig:intersection-icyroad}). Depending on the episode the oncoming car will randomly turn left or right with $50\%$ chance. If it turns left, the agent's car must slow down to avoid collision. The observation is $10$ dimensional and contains the positions and velocities of both cars as well as their heading (angle). The agent's car moves in the $+y$ direction, and has $3$ discrete actions for accelerating, decelerating, or neither along the $y$-axis. The reward is:
\begin{equation}
    r(s,a)=\dot{y}_{\text{agent}} - 10 \cdot \delta[\text{collision}]
\end{equation}
where $\delta[\text{condition}]$ is $1$ if the condition is true and $0$ otherwise. The cars are initialized in fixed positions entering the intersection from opposite directions, with a velocity of $5$ (for the agent) or $-5$ (for the oncoming car), and each episode is $25$ timesteps.

\end{document}